# Global Norm-Aware Pooling for Pose-Robust Face Recognition at Low False Positive Rate

Sheng Chen[a,b*], Jia Guo[c], Yang Liu[b], Xiang Gao[b], Zhen Han[a]

[a] School of Computer and Information Technology, Beijing Jiaotong University, Beijing, China
{shengchen, zhan}@bjtu.edu.cn
[b] Research Institute, Watchdata Inc., Beijing, China
{yang.liu.yj, xiang.gao}@watchdata.com
[c] DeepInSight, China
guojia@gmail.com

**Abstract**

In this paper, we propose a novel Global Norm-Aware Pooling (GNAP) block, which reweights local features in a convolutional neural network (CNN) adaptively according to their L2 norms and outputs a global feature vector with a global average pooling layer. Our GNAP block is designed to give dynamic weights to local features in different spatial positions without losing spatial symmetry. We use a GNAP block in a face feature embedding CNN to produce discriminative face feature vectors for pose-robust face recognition. The GNAP block is of very cheap computational cost, but it is very powerful for frontal-profile face recognition. Under the CFP frontal-profile protocol, the GNAP block can not only reduce EER dramatically but also boost TPR@FPR=0.1% (TPR i.e. True Positive Rate, FPR i.e. False Positive Rate) substantially. Our experiments show that the GNAP block greatly promotes pose-robust face recognition over the base model especially at low false positive rate.

## 1. Introduction

Face recognition performance has been substantially promoted by deep learning for the last several years. Many studies focus on face recognition under constrained conditions. Despite the success of near-frontal face recognition, large pose variation is still a big challenge for unconstrained face recognition. Many face recognition algorithms' performance drops suddenly by over 10% when it comes from frontal-frontal to frontal-profile face verification [1].

Recently, several pose-robust face recognition algorithms have been proposed, for example, [2, 3, 4]. Under the CFP frontal-profile (CFP-FP) protocol [1], previous state-of-the-art pose-robust face recognition algorithms can reduce equal error rate (EER) by 16.3% ~ 23.7%. However, all of them add too much complexity to the base models or the training process.

---

* Corresponding author. E-mail address: shengchen@bjtu.edu.cn

In this paper, we propose a novel Global Norm-Aware Pooling (GNAP) block, which reweights local features in a convolutional neural network (CNN) adaptively according to their L2 norms and outputs a global feature vector with a global average pooling layer. We use a GNAP block in a face feature embedding CNN to produce discriminative face feature vectors for pose-robust face recognition. The GNAP block is of very cheap computational cost, but it is very powerful for frontal-profile face recognition. Under the CFP frontal-profile protocol, the GNAP block can not only reduce EER dramatically but also boost TPR@FPR=0.1% (TPR i.e. True Positive Rate, FPR i.e. False Positive Rate) substantially.

The GNAP block is inspired by the observation that the spatial symmetry of the global operator in a face feature embedding CNN is very helpful to the pose robustness of the face feature vectors. Both global average pooling (GAP) layer and fully connected (FC) layer are popular global operators in CNNs. GAP is of full spatial symmetry, but FC is usually of no spatial symmetry. We find that GAP can produce pose-robust face feature vector in the sense of reducing EER of CFP-FP compared with FC (see Table 1). However, at very low false positive rate (e.g. FPR=0.1%) on CFP-FP, the face CNN model with GAP achieves much lower TPR than that with FC. Without spatial symmetry, FC gives different weights to different units of its input, which is quite different from GAP.

Our GNAP block is designed to give dynamic weights to local features in different spatial positions without losing spatial symmetry. It is an ingenious way to reweight the local features adaptively by according to their L2 norms, which gives different weights in different spatial positions and keeps spatial symmetry. The GNAP block also uses global average pooling layer to do down-sampling in a fully symmetrical way.

Compared with FC, GAP and several variants of the GNAP block, our GNAP block achieves significantly improved TPR@FPR=0.1% on CFP frontal-profile benchmark under the same experimental conditions. We also give another explanation for why the GNAP block performs the best.

The novelties of our GNAP block are summarized as follows: (1) Implementing GNAP block is very easy in the popular deep learning frameworks such as Caffe, Tensorflow, Mxnet, and Pytorch. Moreover, the GNAP block adds no complexity to the training process. (2) The GNAP block is very lightweight. Except for the few parameters of batch normalization (BN, c.f. [9]) layer, the GNAP block adds no parameters and negligible computation to the base model. (3) The GNAP block greatly promotes pose-robust face recognition over the base model especially at low false positive rate (see Table 1).

## 2. Related Work

Recently many researchers have made a try to solve the pose-robust face recognition problem. UV-GAN [3] devises a meticulously designed architecture that combines local and global adversarial DCNNs to learn an identity-preserving facial UV completion model. By combining pose augmentation during training and pose discrepancy reduction during testing, UV-GAN [3] achieves 94.05% verification accuracy under the CFP frontal-profile protocol only. For face recognition in the wild, [2] proposes a Pose Invariant Model (PIM), containing a Face Frontalization sub-Net (FFN) and a Discriminative Learning sub-Net (DLN), which are jointly learned from end to end. PIM [2] with base model Light CNN-29 [8] achieves 93.1% verification accuracy under the CFP frontal-profile protocol. [4] hypothesizes that there is an inherent mapping between frontal and profile faces, and consequently, their discrepancy in the deep representation space can be bridged by an equivariant mapping. To exploit this mapping, a novel Deep Residual EquivAriant Mapping (DREAM) block is formulated in [4], which is capable of adaptively adding residuals to the input deep representation to transform a profile face representation to a canonical pose that simplifies recognition. The DREAM block [4] reduces the error of ResNet-18 and ResNet-50 by 16.3% and 23.7% on the CFP benchmark, achieving 7.03% and 6.02 EER, respectively. [4] does not utilize Generative Adversarial Network (GAN), while [2] and [3] do.

Most studies pose-robust face recognition only report their average accuracy or EERs on cross pose face recognition benchmarks. However, scarce studies have payed attention to their accuracy at low false positive rate on these benchmarks. In this paper, we demonstrate that some models can achieve higher average accuracy but cannot achieve higher TPR at low FPR on the same cross pose face recognition benchmark. Our GNAP block can kill two birds with one stone. Our work is motivated by DREAM block in [4]. However, the model with our GNAP block is much simpler to implement and much easier to train than that with DREAM block.

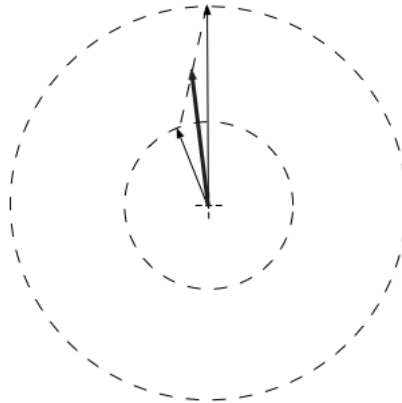

**Fig. 1.** GAP makes the global feature vector overfit the local features with easy viewpoint (that with much larger L2 norm).

### 3. Global Norm-Aware Pooling Block

We first make an analysis on the advantage of spatial symmetry and spatial reweighting when mapping the last non-trivial feature map to a global feature for pose-robust face recognition. Then, we give the exact definition of the Global Norm-Aware Pooling (GNAP) block, which can satisfy both of the two conditions. We also explain additional advantage that the GNAP block has. Several variants of GNAP are also considered.

### 3.1 The advantage of spatial symmetry and spatial reweighting

When detecting the face landmarks on a large pose face image, we probably get more inaccurate positions than on a frontal face image. The face aligned according to the inaccurate positions of the landmarks will not be so good for face recognition. For a large pose face image, we suppose the aligned face as the input of CNN model has a shift of several pixels from the right position. Moreover, assume the last non-trivial feature map (denoted as FMap-end) that the CNN model output has a shift of one unit from its right position. If the FC layer is used to map FMap-end to a global feature vector, each unit of FMap-end will be given a wrong weight, since the FC layer does not keep spatial symmetry. The spatial symmetry can help the model recognize the misaligned face.

If we use the GAP layer to map FMap-end to a global feature vector with spatial symmetry kept, another problem will be raised, as shown in Fig. 1. Some local features in FMap-end have much larger L2 norms than others. The former ones are called local features with easy viewpoint and the latter ones are called local features with hard viewpoint. The direction of the global feature vector is mainly dominated by the local features with easy viewpoint. Alternately speaking, it overfits the local features with easy viewpoint. To reduce the overfit, we need reweight all the local features. Can we make the reweighting without losing spatial symmetry? The answer is yes, as presented in the following section.

**3.2 The GNAP block**

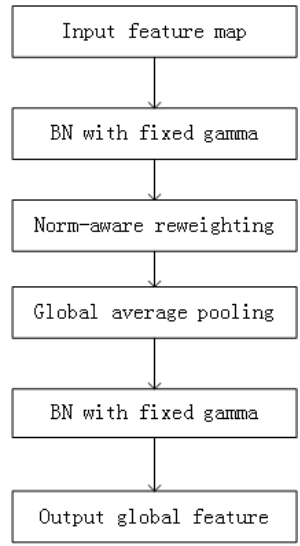

**Fig. 2.** A GNAP block. A GNAP block stacks four layers, each of which keeps spatial symmetry.

To reweight local features without losing spatial symmetry, we present a GNAP block, which can also solve other problems and promote the robustness of the global feature for cross pose face recognition. As shown in Fig. 2, a GNAP block stacks two BN (with fixed gamma, c.f. Mxnet [10]) layers (one layer at the beginning and one layer at the end), a norm-aware reweighting layer, and a GAP layer. In the GNAP block, batch normalization can give different weights to different channels of the input without losing spatial symmetry. BN with fixed gamma can make each channel invariant to the scale change and the basis shift.

Let $F$ be a feature map of spatial resolution $W \times H$ and of channel number $C$. For a fixed spatial position $(i,j)$, the vector $F_{i,j}$ consisted of $F_{i,j,c}$ in all channels is called a local feature of $F$. The L2 norm of the local feature $F_{i,j}$ is computed as:

$$\|F_{i,j}\| = \sqrt{\sum_c F_{i,j,c} \cdot F_{i,j,c}} \qquad (1)$$

The mean of all local features' L2 norms is computed as:

$$\|F\|_{mean} = \frac{1}{W \cdot H} \sum_{i,j} \|F_{i,j}\| \qquad (2)$$

The output for the norm-aware reweighting layer is computed as:

$$G_{i,j,c} = \frac{\|F\|_{mean}}{\|F_{i,j}\|} \cdot F_{i,j,c} \qquad (3)$$

The reweighting ratio in (3) is also invariant to the whole scale change of FMap-end.

If we want to change the channel number of the input or output, we can add pointwise convolution (cov1x1) layers before or after the GNAP block. Pointwise convolution can also give different weights to different channels of the input without losing spatial symmetry.

## 4. Experiments

Following the experimental settings in [5] and [6], we use ArcFace loss to train all face models on CASIA-Webface [7] datasets. We use LResNet18E and MobileFaceNet as our baseline models.

Let FasterFC denote an alternative hyperparameter setting as follows. Set the learning rate of the last FC layer (that is not included in the feature embedding CNN) to be 10x over the learning rate of the other layers. And set all the weight decay parameters to be 5e-4.

Let LResNet18E(GAP) denote the LResNet18E variant in which GAP is used to replace the penultimate FC layer. Let LResNet18E(GNAP) denote the LResNet18E variant in which GNAP is used to replace the penultimate FC layer. LResNet18E(FasterFC), LResNet18E(GAP), and LResNet18E(GNAP) are trained with FasterFC setting. We compare their performance in Table 1. The GNAP block with FasterFC hyperparameter setting greatly promotes the pose-robustness of LResNet18E especially at low FPR.

**Table 1.** Performance comparison among face models trained on CASIA-Webface. "TPR" refers to face verification TPR under FPR = 0.1%.

| Network | Acc(LFW) | TPR(LFW) | Acc(CFP-FP) | EER(CFP-FP) | TPR(CFP-FP) |
|---|---|---|---|---|---|
| LResNet18E | 99.23% | 97.66% | 94.43% | 6.14% | 75.31% |
| LResNet18E(FasterFC) | 99.32% | 97.63% | 94.97% | 5.06% | 76.07% |
| LResNet18E(GAP) | 99.17% | 97.53% | 94.80% | 5.54% | 74.69% |
| **LResNet18E(GNAP)** | **99.33%** | **98.27%** | **95.19%** | **4.97%** | **78.31%** |

Let MobileFaceNet+ denote the MobileFaceNet variant in which GNAP is used to replace the global depthwise convolution layer. And let MobileFaceNet++ denote the MobileFaceNet+ variant trained with FasterFC setting. Their performance on CFP-FP are compared in Table 2. The GNAP block greatly promotes the pose-robustness of MobileFaceNet.

**Table 2.** GNAP greatly promotes the pose-robustness of MobileFaceNet when trained on CASIA-Webface. "TPR" refers to face verification TPR under FPR = 0.1%.

| Network | Acc(CFP-FP) | TPR(CFP-FP) |
|---|---|---|
| MobileFaceNet | 88.97% | 67.63% |
| MobileFaceNet+ | 94.43% | 75.43% |
| **MobileFaceNet++** | **94.95%** | **79.49%** |

## 5. Conclusion

The GNAP block can reweight local features without losing spatial symmetry. It is of very cheap computational cost but very powerful for frontal-profile face recognition. Under the CFP frontal-profile protocol, the GNAP block can not only reduce EER dramatically but also boost TPR@FPR=0.1% substantially.